\documentclass[runningheads]{llncs}
\usepackage[T1]{fontenc}
\usepackage{graphicx}
\usepackage{booktabs}
\usepackage{adjustbox}
\usepackage{multirow}
\usepackage{subcaption}
\usepackage{comment}
\usepackage{siunitx}
\usepackage{xcolor}
\usepackage{enumitem}
\usepackage{tabularx}
\usepackage{textcomp}

\begin{document}
\title{A Generative Approach for Image Registration of Visible-Thermal (VT) Cancer Faces}
%
%
\author{Catherine Ordun\inst{1,2} \and
Alexandra Cha\inst{1} \and
Edward Raff\inst{1,2} \and
Sanjay Purushotham\inst{1,2} \and
Karen Kwok\inst{3} \and
Mason Rule \inst{3}\and
James Gulley \inst{3}
}
\authorrunning{C. Ordun et al.}
%
\institute{Booz Allen Hamilton, Washington, DC USA \\
\email{\{cha\_alexandra, raff\_edward\}@bah.com} \and
University of Maryland Baltimore County, Baltimore, MD USA \\
\email{\{cordun1, psanjay\}@umbc.edu} \and
National Cancer Institute, National Institutes of Health, Rockville, MD USA \\
\email{\{karen.kwok, mason.rule, gulleyj\}@nih.gov}
}
\maketitle              
\begin{abstract}
Since thermal imagery offers a unique modality to investigate pain, the U.S. National Institutes of Health (NIH) has collected a large and diverse set of cancer patient facial thermograms for AI-based pain research. However, differing angles from camera capture between thermal and visible sensors has led to misalignment between Visible-Thermal (VT) images. We modernize the classic computer vision task of image registration by applying and modifying a generative alignment algorithm to register VT cancer faces, without the need for a reference or alignment parameters. By registering VT faces, we demonstrate that the quality of thermal images produced in the generative AI downstream task of Visible-to-Thermal (V2T) image translation significantly improves up to 52.5\%, than without registration. \emph{Images in this paper have been approved by the NIH NCI for public dissemination.}

\keywords{Chronic Pain \and Generative AI \and Generative Adversarial Network \and Diffusion Models \and Visible-Thermal Image Translation}
\end{abstract}

\section{Introduction}
\begin{figure*}[t]
\centering
     \begin{subfigure}[b]{0.49\textwidth}
         \includegraphics[width=\textwidth]{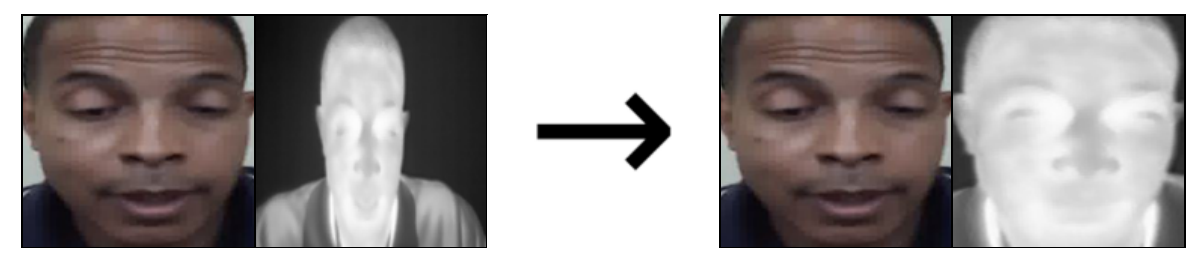}
     \end{subfigure} 
     \begin{subfigure}[b]{0.49\textwidth}
         \includegraphics[width=\textwidth]{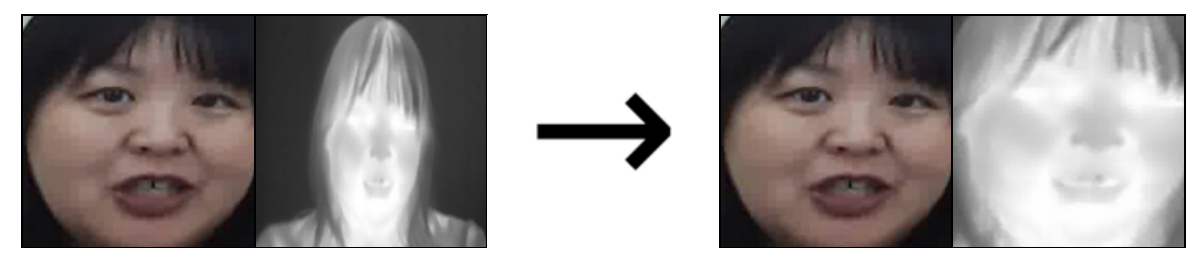}
     \end{subfigure} 
     \begin{subfigure}[b]{0.49\textwidth}
         \includegraphics[width=\textwidth]{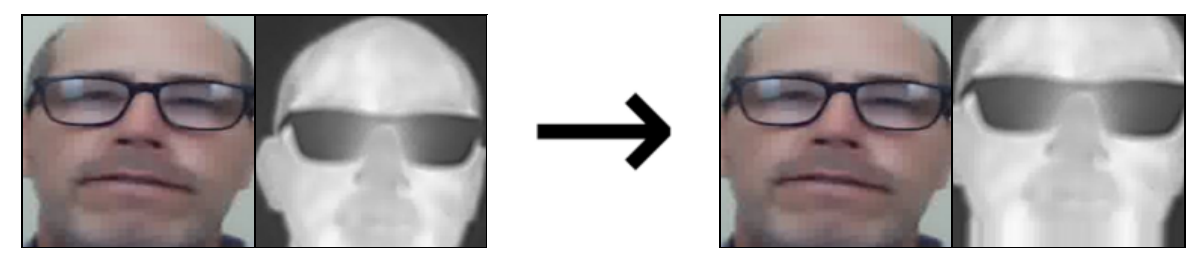}
     \end{subfigure} 
    \begin{subfigure}[b]{0.49\textwidth}
         \includegraphics[width=\textwidth]{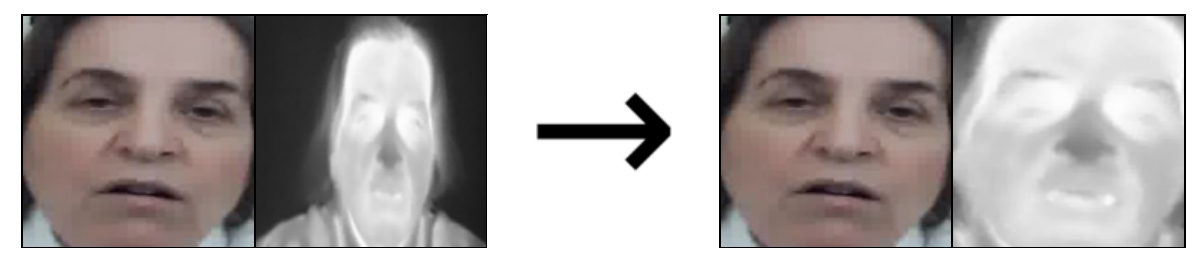}
     \end{subfigure} 
     \begin{subfigure}[b]{0.49\textwidth}
         \includegraphics[width=\textwidth]{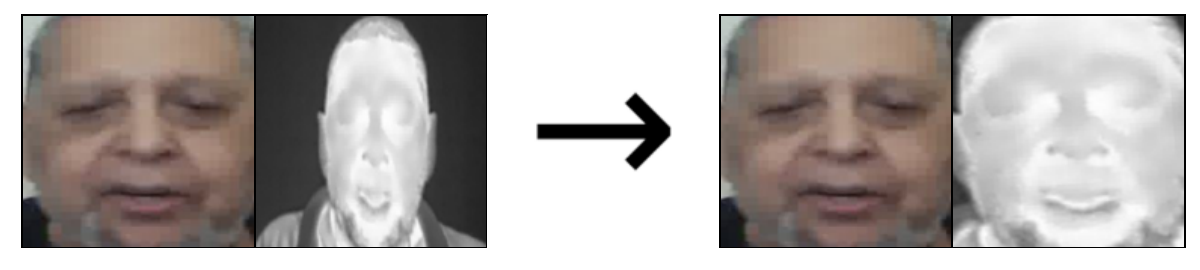}
     \end{subfigure} 
     \begin{subfigure}[b]{0.49\textwidth} 
         \includegraphics[width=\textwidth]{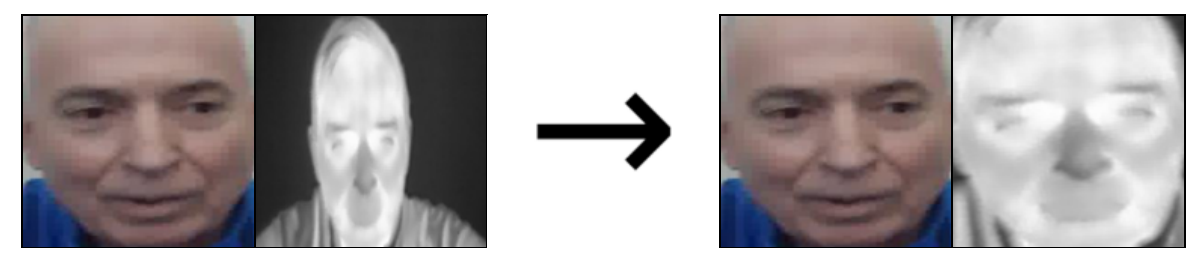}
     \end{subfigure} 
    \caption{ISS VT Facial Pairs Before and After Alignment using a Generative Approach for Image Registration for Six Patients.}
    \label{marquis}
\end{figure*}

Thermal temperatures have been shown to directly correlate with gold standard vital metrics that measure physiological excitement such as electrocardiography (ECG) and galvanic skin response (GSR), making it a promising non-invasive measure for pain assessment \cite{ioannou2014thermal,ordun2020use,pavlidis2000imaging,puri2005stresscam}. Further, thermograms reveal signs of pain by detecting temperature variations in inflamed tissue such as heat patterns in the supraorbital and periorbital facial areas at the onset of a migraine attack \cite{pavlidis2018dynamic}, and mean temperature elevation in swollen and tender body regions when analyzing arthritis, tennis elbow, and fibromyalgia \cite{ring2012infrared}. As a result, the Intelligent Sight and Sound (ISS) \cite{ordun2022intelligent} study under the U.S. National Institutes of Health (NIH), National Cancer Institute (NCI), has collected thermal images of cancer patient faces since October 2020. The study's objective is to assess the use of AI as a tool to detect chronic cancer pain - an understudied problem in machine learning and thermal physiology. For many AI tasks, such as generative modeling, the thermal image must be paired with its corresponding visible image in order to extract multimodal signals. Further, the pairing must be exact since shifts in pixels can lead to significant changes in model prediction and generative AI image quality \cite{kong2021breaking,zhang2019making}. Multi-spectral image alignment between thermal and visible images is a non-trivial task, that is further exacerbated when a reference for scale, such as deformation parameters, is not available. To address this problem, we train and modify Vista Morph \cite{ordun2023vista}, a generative multi-spectral image registration framework, on the ISS VT Facial Dataset of cancer patients. Vista Morph predicts the affine matrix parameters used to align thousands of thermal images, respective to the geometry of their visible pairs, even under extremely warped conditions, leading to registered VT facial pairs shown in Figure \ref{marquis}. In addition, automatic registration saves significant human labor otherwise spent manually aligning images by hand, or relying on proprietary, thermal landmark libraries. This paper offers multiple contributions: 1) We introduce the ISS VT Facial Dataset, which is to the best of our knowledge, the largest cancer VT facial dataset available, useful for thermal physiology, thermal facial recognition, and generative AI studies consisting of 29,461 VT facial pairs. 2) We train and modify the generative Vista Morph image registration algorithm to correct for challenging, severe misalignment between VT faces.  3) We show that registered images generate improved quality of thermal faces than without registration using GANs, as a method of synthetic data augmentation. 

\section{Methodology}
\subsection{ISS VT Facial Dataset}
\textbf{Capture Protocol.}
VT facial images are extracted from videos of cancer patients from the Intelligent Sight and Sound (ISS) Dataset \cite{ordun2022intelligent} - an ongoing, observational, non-interventional clinical study initiated by the U.S. National Institutes of Health (NIH) National Cancer Institute (NCI). The study is entitled \emph{``A Feasibility Study Investigating the Use of Machine Learning to Analyze Facial Imaging, Voice and Spoken Language for the Capture and Classification of Cancer Pain}" \cite{NIH}. 

\begin{figure*}[h]
\centering
    \begin{subfigure}[b]{0.21\textwidth}
        \includegraphics[width=\textwidth]{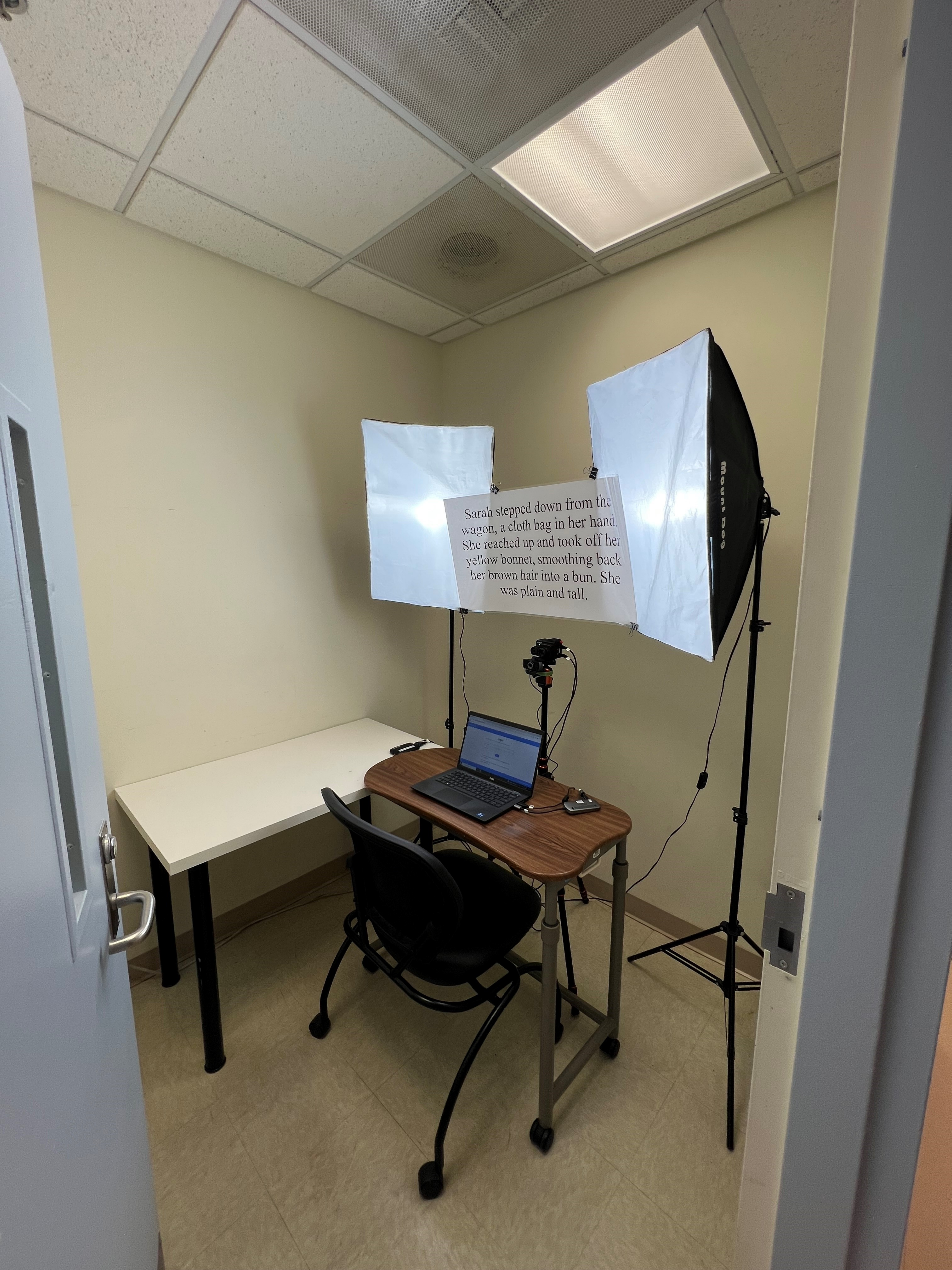}
        \caption{Side View}
     \end{subfigure}    
     \begin{subfigure}[b]{0.21\textwidth}
         \includegraphics[width=\textwidth]{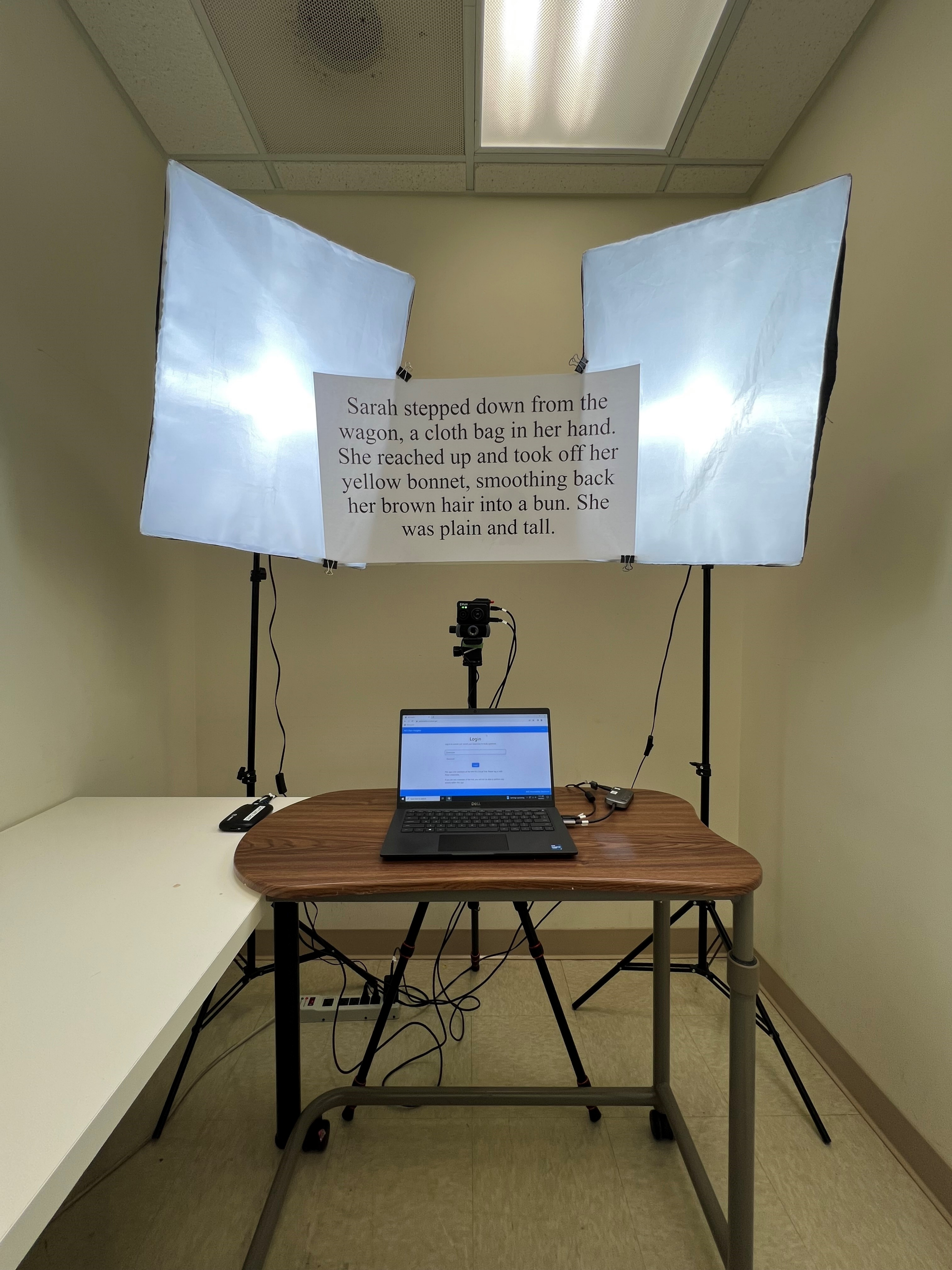}
         \caption{Forward View}
     \end{subfigure} 
    \caption{Experimental Setup to collect VT facial pairs at the NIH clinic.}
    \label{setup}
\end{figure*}

Videos of participant faces are captured both remotely, using a smartphone, and in the NIH clinic.  Smartphone video provides visible spectrum data while the VT pairs are only captured at the NIH clinic. In-clinic, two separate cameras capture video in the visible (0.4-0.7 $\mu m$, NexiGo PC Webcam) and thermal (7.5-13.5 $\mu m$, Flir Duo Pro R) spectra as shown in Figure \ref{setup}. Both cameras are positioned in a vertical stack, to minimize parallax effects, and erected on a tripod less than two feet away from the patient in a frontal position. Only the patient and the cameras are present in the room, no other personnel are present during recording. The protocol begins when the patient reads a 10-15 second-long passage selected from a grade 3 reading level text, displayed on the white poster in Figure \ref{setup}. This is a neutralizing prompt, and is common practice in mood conditioning trials in order to control for an emotionally charged response \cite{apolinario2018improving,fink2020interpretation,livesay1994emg}. Next, the patient records a video response to the prompt \emph{"Please describe how you feel right now,”} designed to capture narratives about the patient's mood, beliefs and attitudes about their pain, medical conditions and daily activities. 

\textbf{Data Processing.}
From 96 VT video recordings, we extract 29,461 VT pairs across 44 subjects using minimal preprocessing. First, we extract frames from VT videos at a rate of 10 fps using the \url{ffmpeg} library. Second, we detect and crop faces using a Multi-task Cascaded Convolutional Network (MTCNN) \cite{zhang2016joint}. Lastly, we apply a series of binary thresholding operations to crop the thermal face from the background using \url{OpenCV}. This crude processing leads to severely misaligned VT facial pairs, as shown in samples per Figure \ref{marquis} before registration. The incidence of older aged individuals who tend to look down while speaking, as opposed to maintaining frontal head alignment, in addition to low resolution from both sensors, and dynamic-recorded video sessions, makes the ISS VT Facial Dataset more challenging for alignment than other VT facial datasets that are captured in standard settings (i.e. background, prompted poses) such as ARL Devcom \cite{poster2021large}, Eurecom \cite{mallat2018benchmark}, and Carl \cite{espinosa2013new} that use similar VoX microbolometer thermal sensors. 

\begin{figure*}[t]
\centering
    \includegraphics[width=0.99\textwidth]{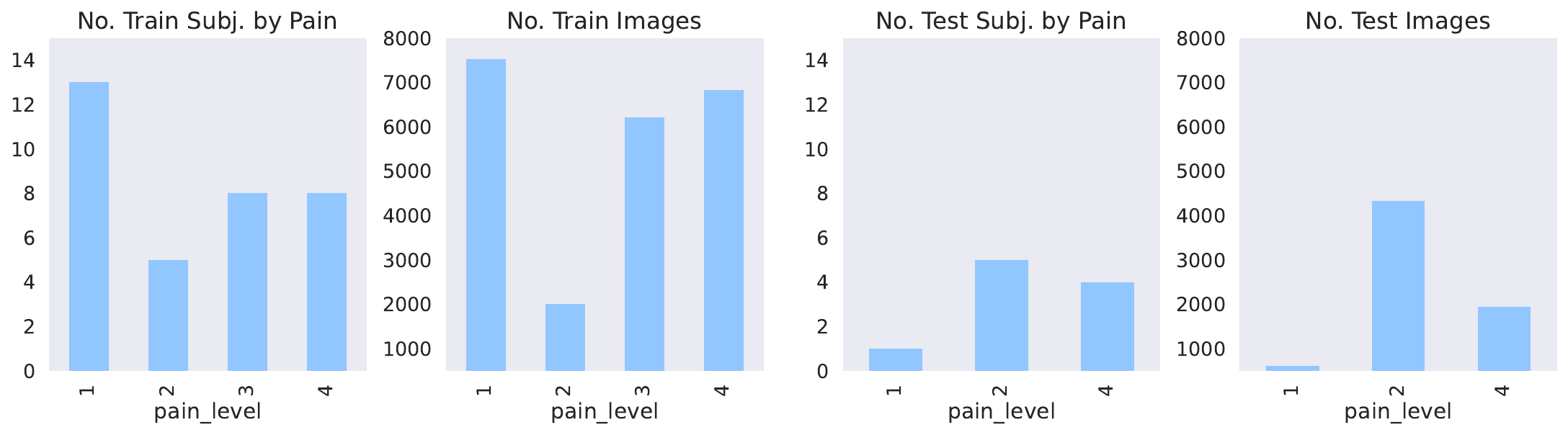}
    \caption{Distribution of Pain Classes in the NIH ISS VT Facial Dataset.}
    \label{stats}
\end{figure*}

\textbf{Data Overview.}
Samples are shown in Figure \ref{marquis}. We split subjects into non-overlapping train and test sets, so that subjects are seen only once in either group. There are 22,570 VT image pairs in the training set and 6,891 VT image pairs in the test set. The training set consists of 34 subjects (67\% male, 33\% female) extracted from 75 videos representing 37.6 minutes of video. Shown in Figure \ref{stats}, approximately 38\% of train subjects have No Pain (1), 14\% have Low Pain (2), 24\% have Moderate Pain (3) and 24\% with Severe Pain (4). The test set consists of 10 test subjects (40\% male, 60\% female) extracted from 21 videos, representing 11.5 minutes of video. In the test set, approximately 50\% of the subjects have Low Pain (2), 40\% have Severe Pain (4), and 10\% have No Pain (1). No test subjects represent Moderate Pain (3) as shown in Figure \ref{stats}. 

\subsection{Vista Morph Framework}
\begin{figure*}[h]
\centering
    \begin{subfigure}[b]{0.45\textwidth}
        \includegraphics[width=\textwidth]{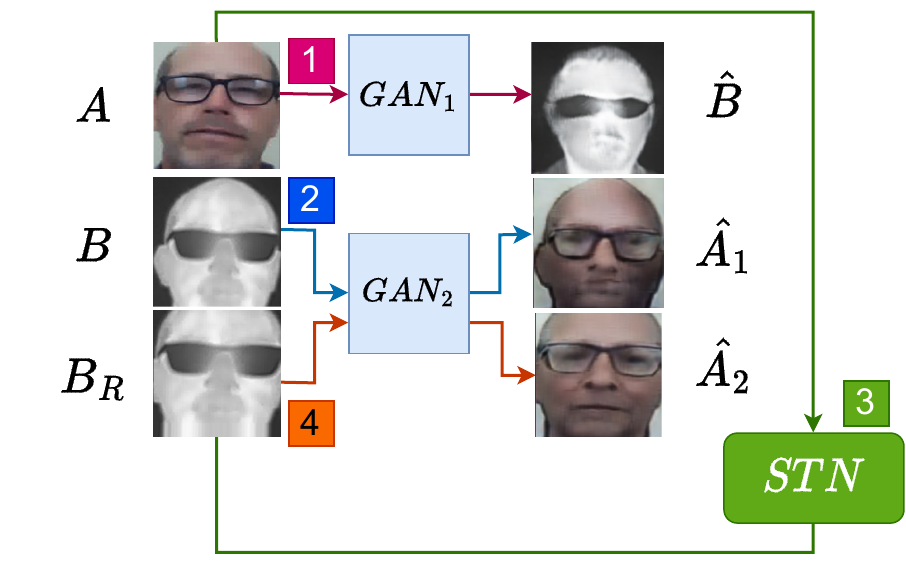}
        \caption{Vista Morph Flows}
        \label{vm_flows}
     \end{subfigure}    
    \begin{subfigure}[b]{0.33\textwidth}
        \includegraphics[width=\textwidth]{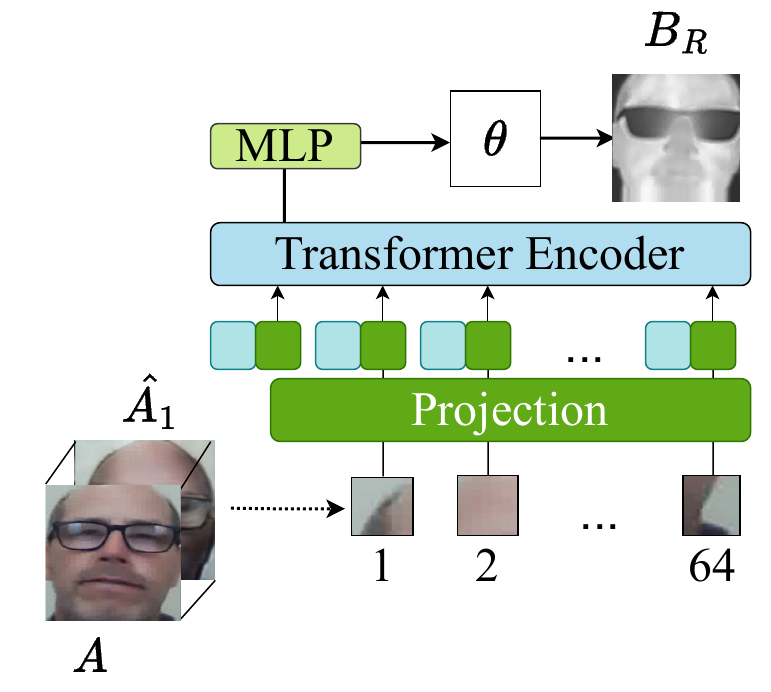}
        \caption{STN Framework}
        \label{stn_arch}
     \end{subfigure}    
     \begin{subfigure}[b]{0.2\textwidth}
         \includegraphics[width=\textwidth]{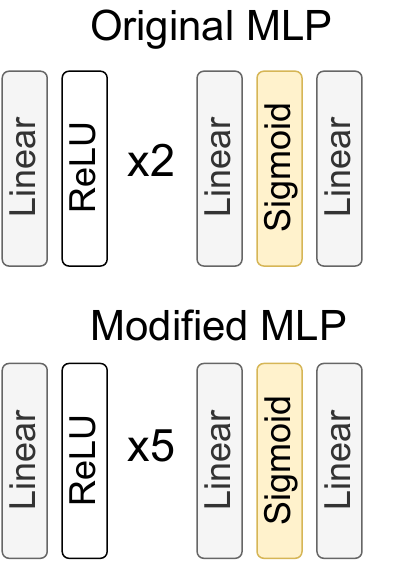}
         \caption{MLP}
         \label{mlp_arch}
     \end{subfigure} 
    \caption{Vista Morph Framework - Training Flows, STN, and MLP Regressor.}
    \label{vistamorph}
\end{figure*}

We train a deep generative image registration algorithm called Vista Morph \cite{ordun2023vista} on the ISS VT Facial Dataset, shown in Figure \ref{vm_flows}. This framework is specifically designed for VT facial pair alignment and is more accurate than comparable algorithms such as Nemar \cite{arar2020unsupervised}. It integrates a Spatial Transformer Network (STN) \cite{jaderberg2015spatial} consisting of a Vision Transformer (ViT) \cite{dosovitskiy2020image}, and two conditional GANs \cite{isola2017image}, and is trained using four flows in an end-to-end manner. The objective is an unsupervised image registration approach. The goal is to ``cast" the misaligned thermal image into a visible spectrum using GANs, in order to learn deformation parameters in a common spectrum (visible) when passed to the STN algorithm. 

Flow 1 consists of a Visible-to-Thermal (V2T) conditional GAN ($GAN_1$) that given, the ground-truth visible input ($A$), translates (generates) its fake thermal ($\hat{B}$) pair. Flow 2 uses a Thermal-to-Visible (T2V) conditional GAN ($GAN_2$) to translate the fake visible image ($\hat{A_1}$) from the thermal ground truth ($B$). Both GANs share the same architecture consisting of a small U-NET \cite{ronneberger2015u} framework. In Flow 3, the STN shown in Figure \ref{stn_arch}, now takes the combination of $(A, \hat{A_1})$ and passes the vector through a ViT as a series of 64 patch embeddings. In the final Flow 4, the same T2V GAN from Flow 2 is used to translate the registered thermal image ($B_R$) to the last intermediate fake thermal image ($A_2$), thereby enforcing a cyclic consistency. Shown in Figure \ref{mlp_arch}, the encoded output is passed to a MLP which acts as a regressor and is used predict the affine matrix ($\theta$) which estimates parameters such as scale, rotation, translation, and shear. We modify the original MLP architecture in Figure \ref{mlp_arch} from two Linear-ReLU blocks to five. Finer control to learn deformation parameters is afforded when the network is deeper to localize geometry of finer features. After predicting the parameters of the affine matrix, the STN samples each pixel in a deformation grid and applies it onto the real thermal image ($B$). The end result is the registered thermal face, $B_R$. For further details regarding the generator and discriminator architectures, training regimen, losses, and ablation studies for robustness, we direct the reader to our previous works in \cite{ordun2023vista,ordun2023visible}. 

\vspace{-1mm}
\subsection{Experiments}
\vspace{-1mm}
\textbf{Image Registration.} The objective of this experiment is to align the thermal facial image with respect to the scale and geometry of the visible face, since no thermal reference is provided. We train the Vista Morph algorithm from scratch, using the entire 22,570 VT pairs, and test on 6,891 pairs. We use the PyTorch library and train on 8 Tesla V100 GPUs in parallel using automatic mixed precision for 10 epochs using a batch\_size=64. To score registration accuracy, we use conventional  metrics for unsupervised image registration - Normalized Cross Correlation (NCC) and Structural Similarity Index Measure (SSIM) \cite{wang2004image} of the edge maps (e.g. morphological gradients of the visible and thermal images), in addition to Mutual Information (MI) \cite{kern2007robust,russakoff2004image}.

\textbf{V2T Image Translation.} The objective of this experiment is to evaluate the quality of generated thermal faces before and after image registration to prove that the quality of generated thermal faces declines without registration leading to artifacts and incorrect identity. We align the entire ISS VT Facial Dataset for all 29,641 pairs using the trained Vista Morph algorithm. We then train a V2T conditional GAN called VTF-GAN \cite{ordun2023visible} on the registered (Vista Morph) and the unregistered (original) training data. We use the PyTorch library and train on 8 Tesla V100 GPUs in parallel using automatic mixed precision for 300 epochs using a batch\_size=64. We use two common generative metrics that measure image quality to score results of the generated thermal face - the Frechet Inception Distance (FID) \cite{heusel2017gans} and LPIPS \cite{zhang2018unreasonable}.

\begin{figure*}[h!]
\centering
     \begin{subfigure}[b]{0.45\textwidth}
         \includegraphics[width=\textwidth]{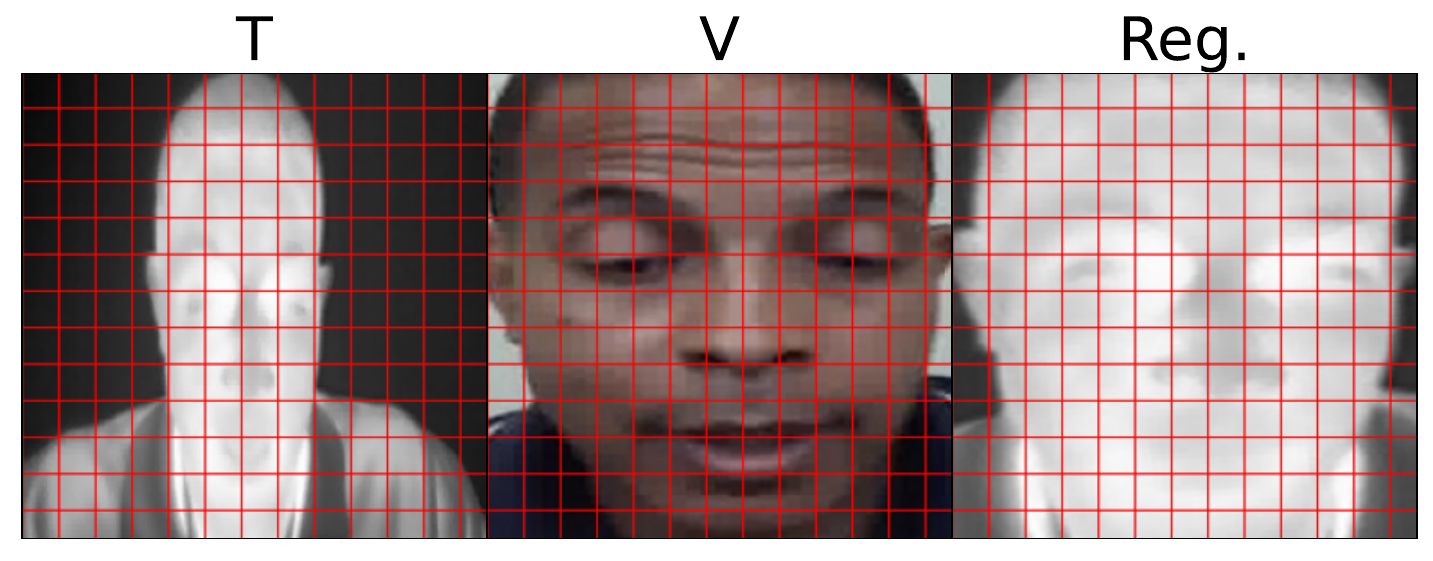}
     \end{subfigure} 
     \begin{subfigure}[b]{0.45\textwidth}
         \includegraphics[width=\textwidth]{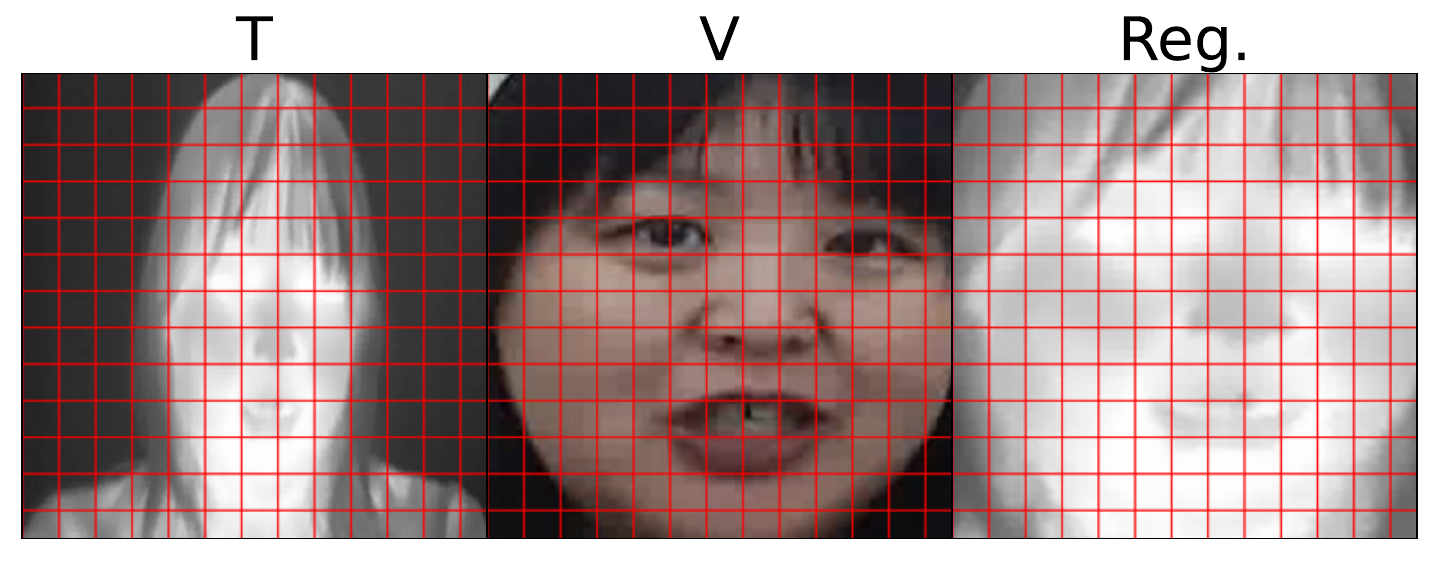}
     \end{subfigure} 
     \begin{subfigure}[b]{0.45\textwidth}
         \includegraphics[width=\textwidth]{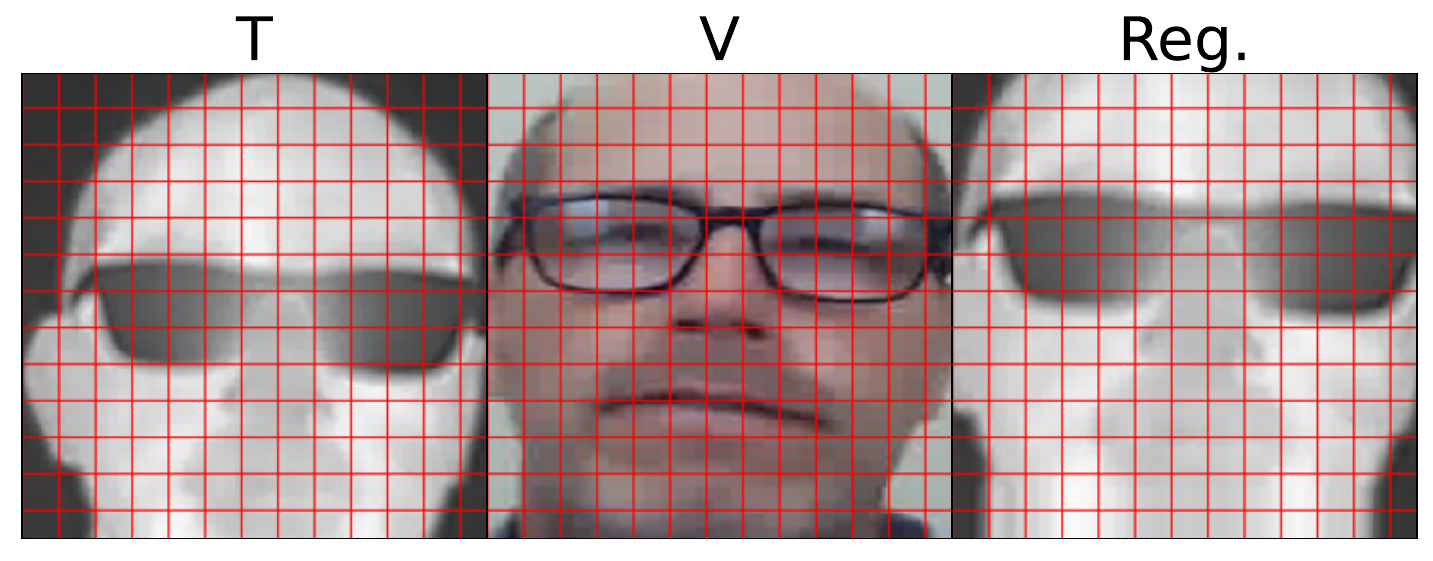}
     \end{subfigure} 
    \begin{subfigure}[b]{0.45\textwidth}
         \includegraphics[width=\textwidth]{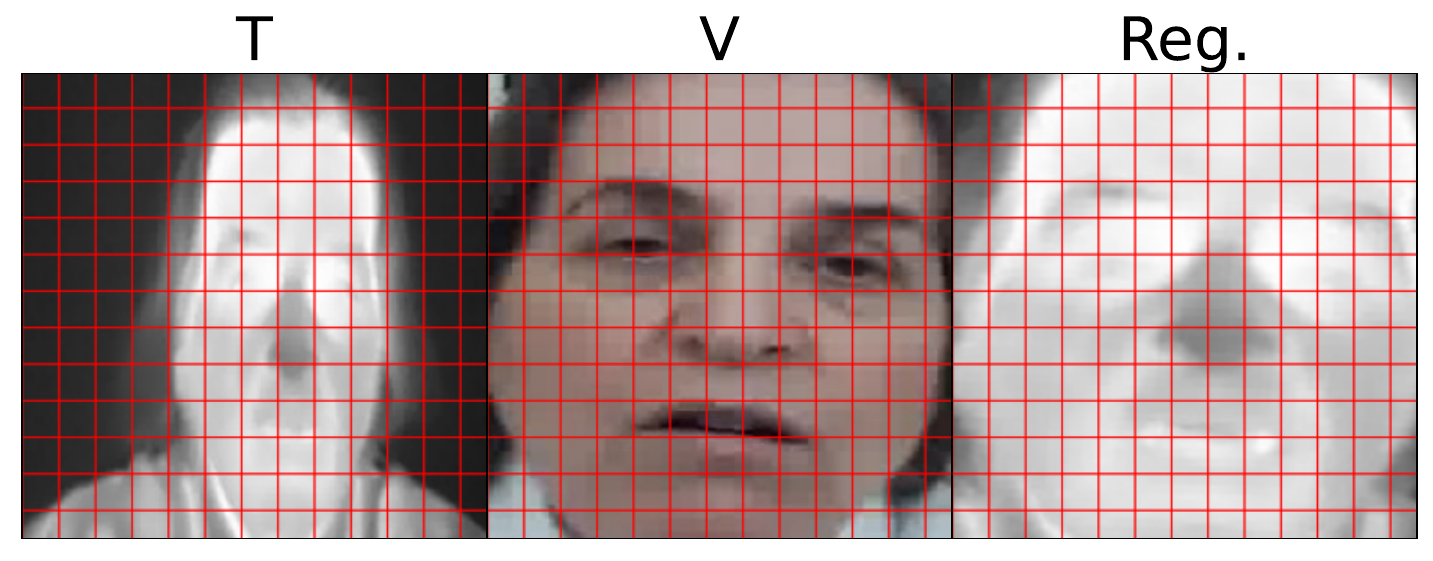}
     \end{subfigure} 
     \begin{subfigure}[b]{0.45\textwidth}
         \includegraphics[width=\textwidth]{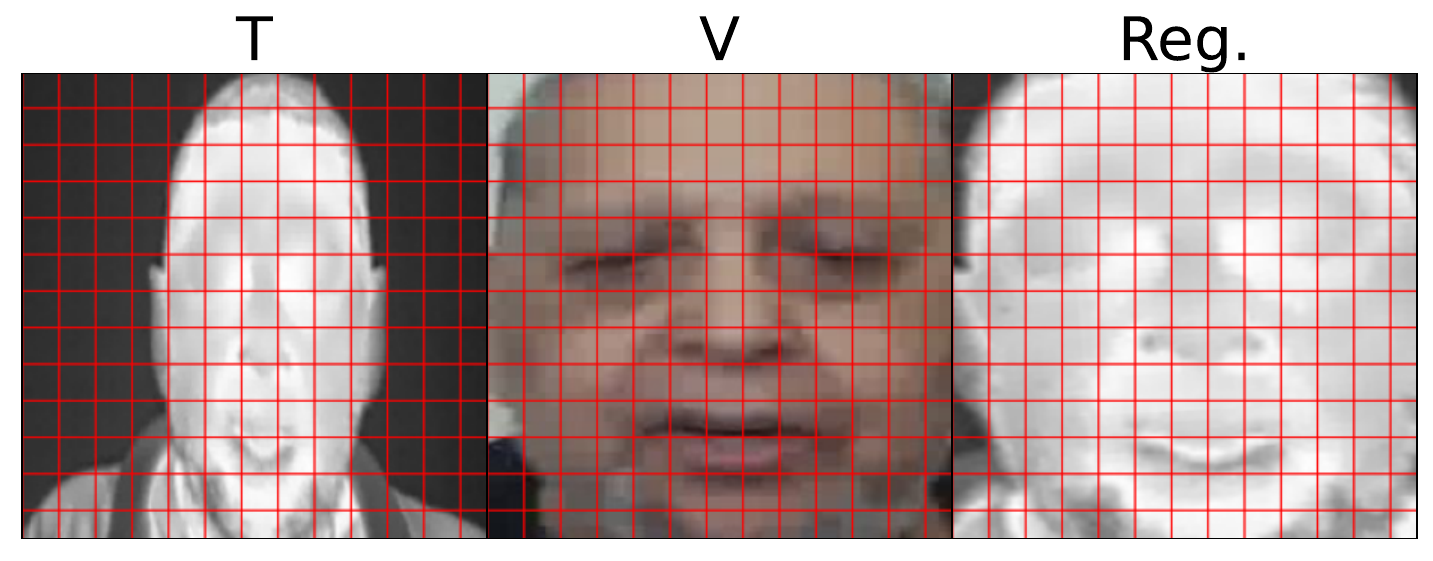}
     \end{subfigure} 
     \begin{subfigure}[b]{0.45\textwidth}
         \includegraphics[width=\textwidth]{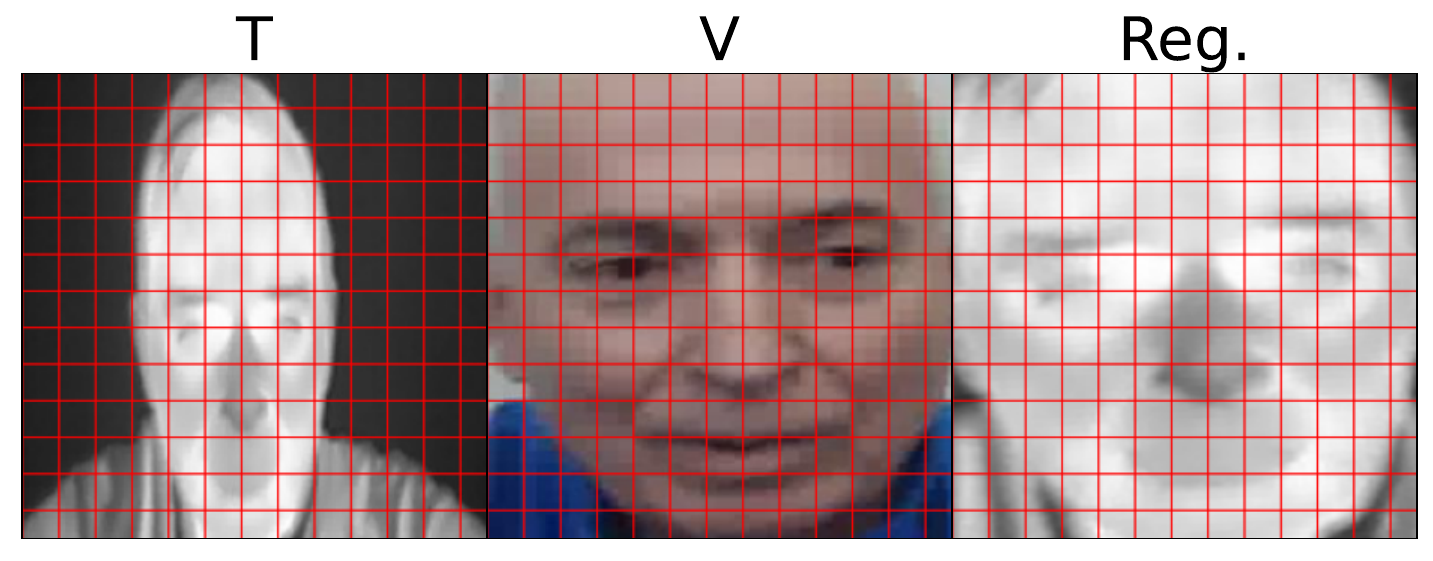}
     \end{subfigure} 
    \caption{Image Registration Samples for Six Patients. }
    \label{registration_samples}
\end{figure*}

\vspace{-10mm}

\begin{table*}[]
\centering
\resizebox{\textwidth}{!}{%
\begin{tabular}{l|ccc|cc}
\toprule
 & \multicolumn{3}{c}{Registration Scores} & \multicolumn{2}{c}{V2T GAN Scores} \\ \addlinespace \midrule
 & SSIM Edges ($\uparrow$) & NCC Edges ($\uparrow$) & Mut. Info. ($\uparrow$) & FID ($\downarrow$) & LPIPS  ($\downarrow$)\\ \midrule
Before Reg. & 0.691 & 0.003 & 0.227 & 121.715 & 0.399 \\ \midrule
After Reg. & \textbf{0.752 (8.9\%)} & \textbf{0.136 (53.4x)} & \textbf{0.299 (31.6\%)} & \textbf{79.680 (-52.8\%)} & \textbf{0.342 (-16.8\%)} \\ \bottomrule
\end{tabular}%
}
\caption{Quantitative Scores for Image Registration and V2T Image Translation.}
\label{registration_scores}
\end{table*}

\vspace{-5mm}
\section{Results}
\begin{figure*}[h]
\centering
     \begin{subfigure}[b]{0.48\textwidth}
         \includegraphics[width=\textwidth]{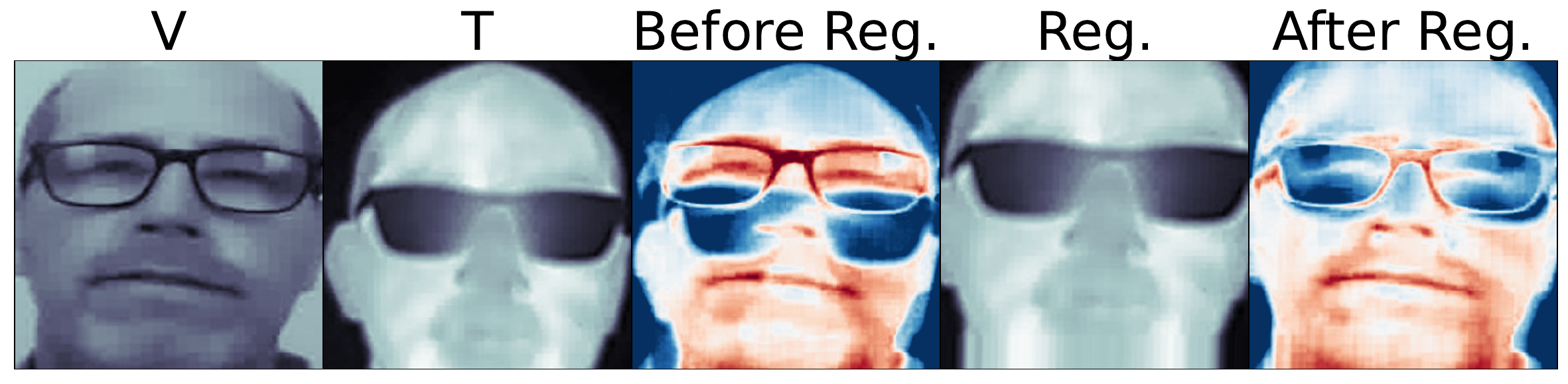}
     \end{subfigure} 
     \begin{subfigure}[b]{0.48\textwidth}
         \includegraphics[width=\textwidth]{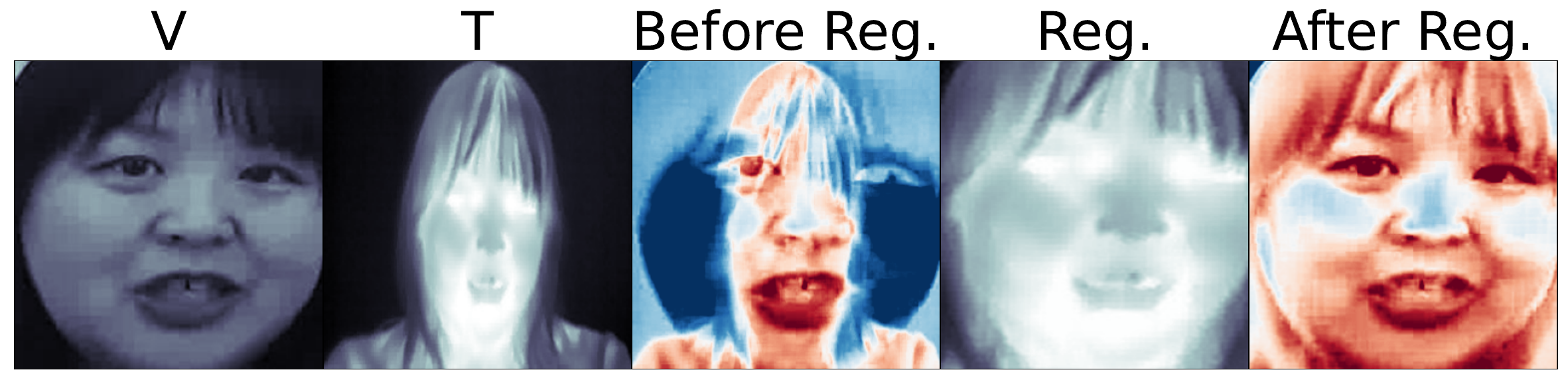}
     \end{subfigure} 
    \caption{Difference Maps Visualizing Alignment Between (V)isible and (T)hermal Pairs Before and After Registration.}
    \label{diffs}
\end{figure*}

\textbf{Image Registration.} Quantitative results shown in Table \ref{registration_scores} demonstrates that alignment of VT faces improve significantly after registration. SSIM scores increase by 8.9\%, NCC increases by 53.4 times, and Mutual Information gains at 31.6\%. Registration samples are shown in Figure \ref{registration_samples} for six test subjects. Grid lines support visual inspection and indicate that the registered thermal face (``Reg.") is well aligned to the scale of the (V)isible ground truth, compared to the original, warped, (T)hermal image. Notice that despite non-frontal head poses such as looking down or upwards in addition to right and left head tilts, that the thermal faces are well aligned. In addition, Figure \ref{marquis} shows that facial obfuscation with eyeglasses does not interfere with registration. Additional evidence is provided in Figure \ref{diffs} that show difference maps between the visible and thermal image before and after registration. Notice that before registration (``Before Reg."), the red (visible) map is obscured under the blue (thermal) map. However, after registration (``After Reg."), the blue and red maps are completely superimposed.

\vspace{-10mm}
\begin{figure*}[ht!]
\centering
     \begin{subfigure}[b]{0.8\textwidth}
         \includegraphics[width=\textwidth]{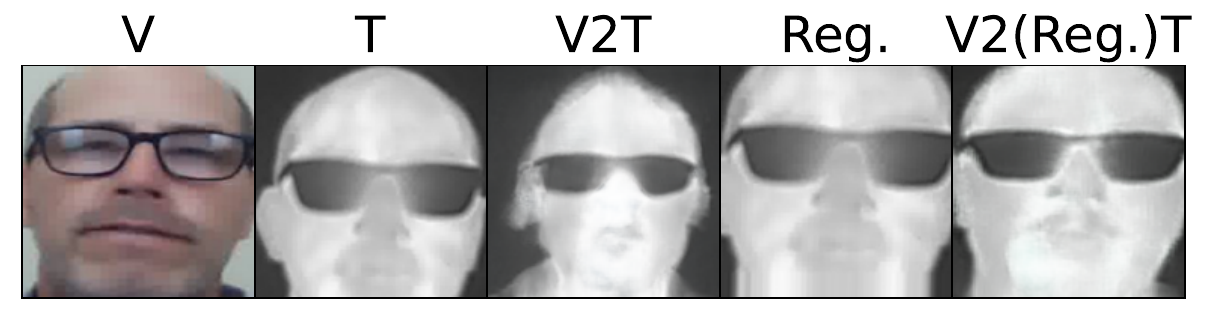}
    \end{subfigure} 
     \begin{subfigure}[b]{0.8\textwidth}
         \includegraphics[width=\textwidth]{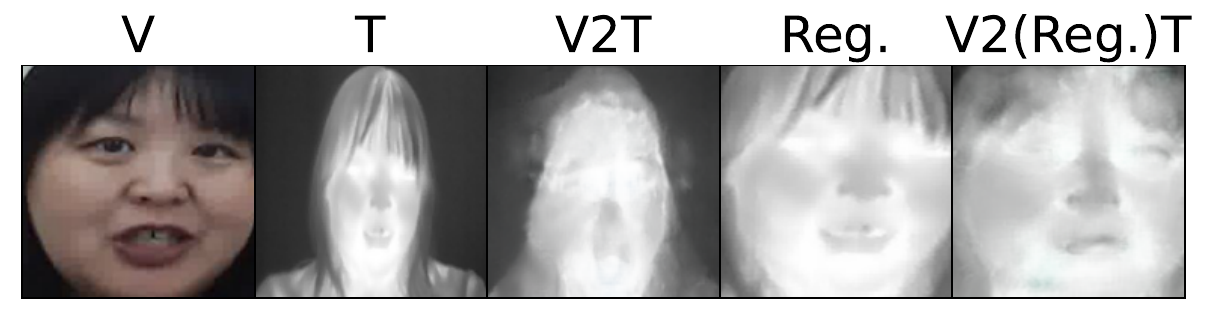}
     \end{subfigure} 
    \caption{V2T Image Translation Samples for Two Patients.}
    \label{gan_results}
\end{figure*}

\textbf{V2T Image Translation.} Table \ref{registration_scores} demonstrates that the quality of thermal faces improves after registration where FID scores improve by 52.8\% and LPIPS scores improve at 16.8\%. In Figure \ref{gan_results}, we show results of the V2T image translation experiments. When training the GAN on the original, misaligned set, the generated thermal is shown in ``V2T". Notice a lack of perceptual clarity, and subjective identity, as well as artifacts. However, when the GAN is trained using the Vista Morph registered VT pairs (``V" and ``Reg."), the ``V2(Reg.)T" generated thermal faces are higher quality and more similar to the original subject.

\textbf{Limitations.}
Image registration is less successful when faces are both highly warped and obfuscated by masks or a combination of glasses and hats. Future works can leverage a Fourier Loss per our work in \cite{ordun2023vista} that is integrated into the Vista Morph framework, used to register VT pairs in No- and Low-Light settings by learning signal frequencies (low, high edges) in addition to the spatial (pixel) domain. Validation in-clinic using a ground-truth thermal sensor should also be conducted to verify accurate heat distribution. 

\begin{figure*}[]
\centering
     \begin{subfigure}[b]{0.48\textwidth}
         \includegraphics[width=\textwidth]{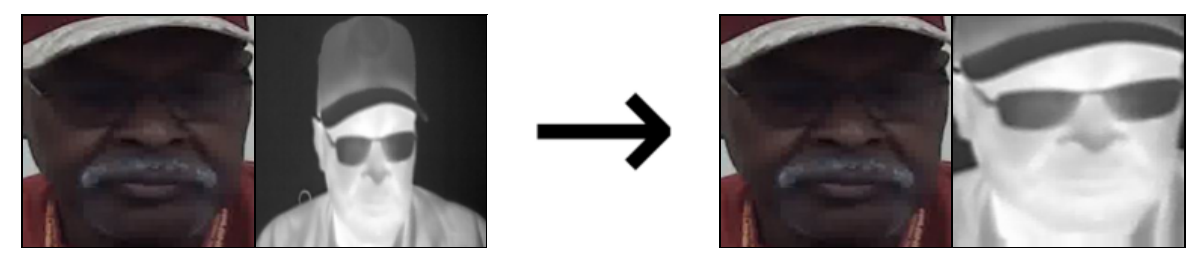}
     \end{subfigure} 
     \begin{subfigure}[b]{.48\textwidth}
         \includegraphics[width=\textwidth]{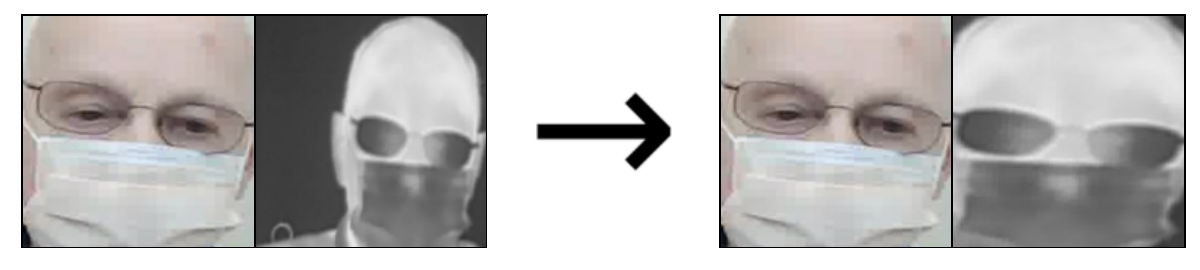}
     \end{subfigure} 
    \caption{Unsuccessful VT Image Registration Samples for Two Subjects.}
    \label{fails}
\end{figure*}

\begin{figure*}[]
\centering
     \begin{subfigure}[b]{0.48\textwidth}
         \includegraphics[width=\textwidth]{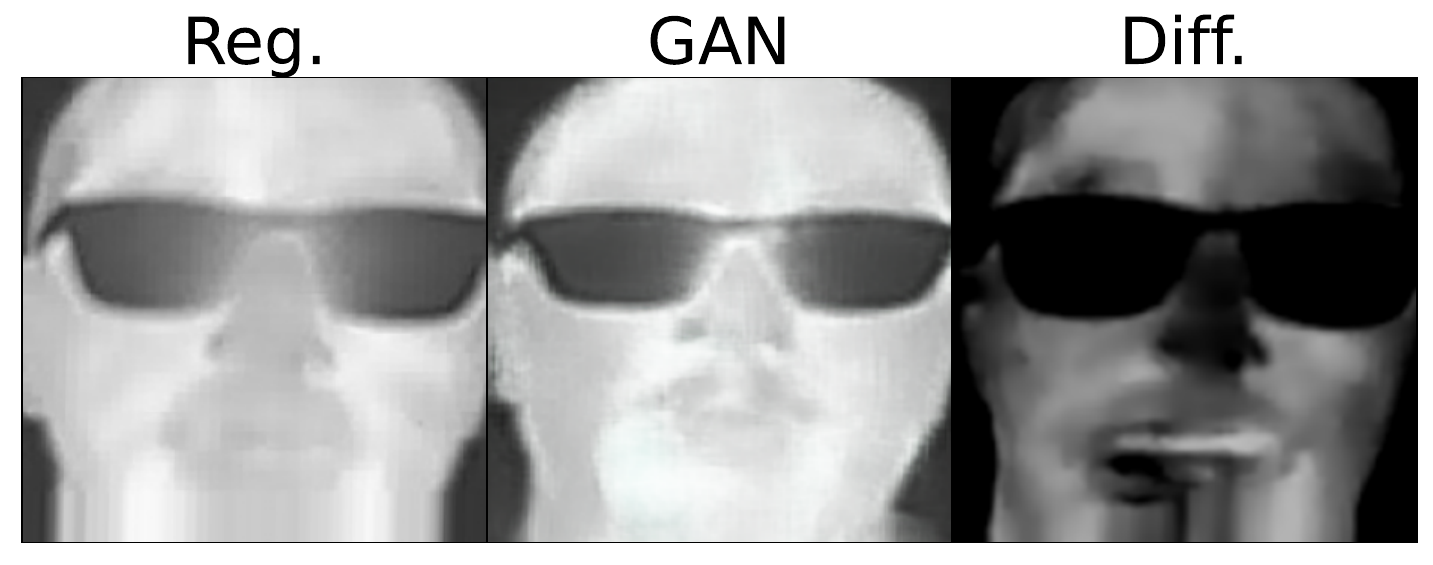}
         \caption{}
     \end{subfigure} 
     \begin{subfigure}[b]{0.48\textwidth}
         \includegraphics[width=\textwidth]{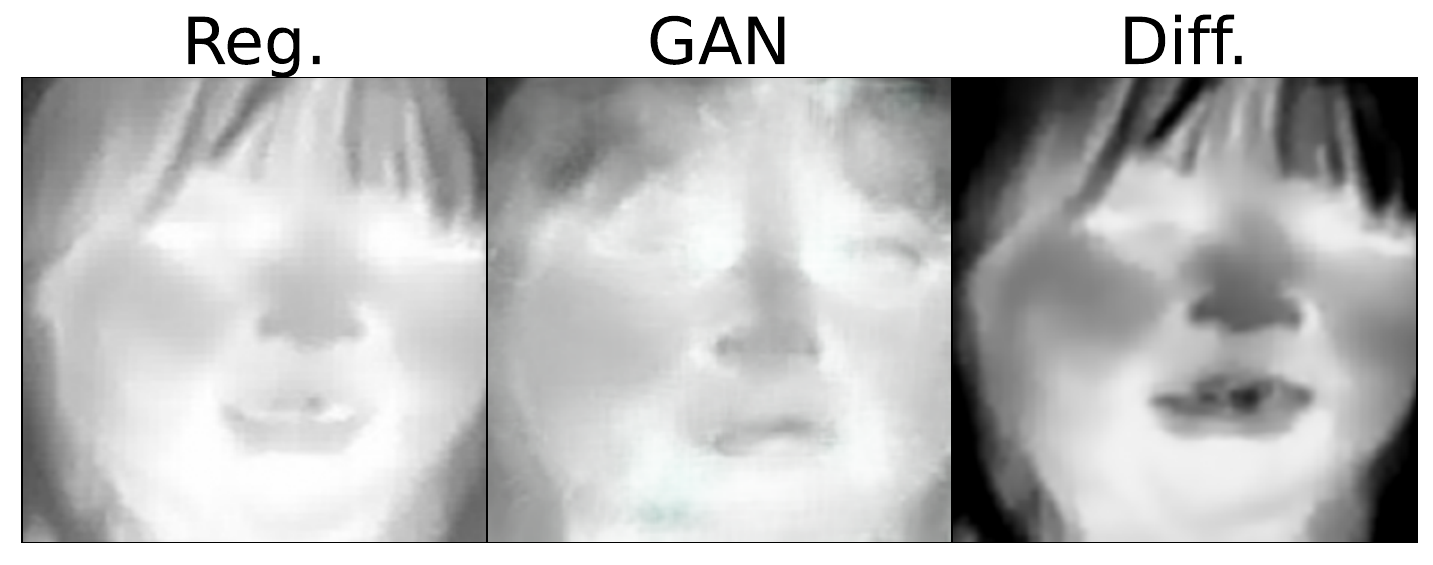}
         \caption{}
     \end{subfigure} 
    \caption{V2T Image Translation with Diffusion Model (``Diff") Leads to Inaccurate Heat Distribution and Artifacts.}
    \label{diffusion}
\end{figure*}

In addition to the VTF-GAN, we also trained a conditional diffusion model called VTF-Diff for V2T image translation \cite{ordun2023visible}. The FID (96.420, +20.8\%) and LPIPS scores (0.440, +22.4\%) were not as competitive as the VTF-GAN results, since most images were distorted. However, we show in Figure \ref{diffusion}, a limited number of successful generated thermal faces (``Diff."). Although these results preserve spatial geometry, the diffusion results are discolored. This implies inaccurate distribution of heat (light pixels) and cold (dark pixels) compared to the ground truth (``Reg."), which can lead to misleading medical assessment. As a result, the VTF-GAN algorithm is preferable over VTF-Diff as a method to augment the ISS VT Facial dataset with additional thermal faces.

\vspace{-3mm}
\section{Conclusion}
We introduce the NIH ISS Visible-Thermal (VT) Facial Dataset, the largest VT cancer dataset to our knowledge, consisting of 29,000 VT pairs for AI pain research. We register all images using a novel, generative approach for multispectral facial imagery and modify the regressor network for finer estimation of alignment parameters. We show that generative tasks such as V2T image translation improve markedly after pairs are registered leading to improved image quality and resolution. Further, our approach demonstrates an effective method for applying two image modalities towards the investigation of cancer pain assessment.

\subsubsection{Acknowledgements} The authors would like to thank Elizabeth Lamping, Katherine Lee-Wisdom, NIH NCI clinic partners (Prostate, Thymoma, Phase 1, Neurofibroma, HIV, GI clinics) for the clinical patient protocol, in addition to Alex Hanson at Booz Allen Hamilton for providing computational support. The authors also thank the patients and their families for informed consent.

%
%
%
\bibliographystyle{splncs04}
\bibliography{references}

\end{document}